%% file: emnlp2024.tex
\pdfoutput=1

\documentclass[11pt]{article}

\usepackage[final]{style/emnlp2023}

\usepackage{times}
\usepackage{latexsym}
\usepackage{tabularx}
\usepackage{makecell}
\usepackage{booktabs}
\usepackage{graphicx}
\usepackage{amsmath}
\usepackage{todonotes}
\usepackage{subcaption}

\PassOptionsToPackage{dvipsnames}{xcolor}

\usepackage{microtype}
\usepackage{float}
\usepackage{inconsolata}
\usepackage[T1]{fontenc}
\usepackage[utf8]{inputenc}

\title{Gradient Localization Improves Lifelong Pretraining of Language Models}
\author{
    Jared Fernandez  \qquad Yonatan Bisk \qquad Emma Strubell
\\ Carnegie Mellon University
\\ \texttt{\{jaredfern, ybisk, strubell\}@cmu.edu}}

\begin{document}
\newcommand{\proposed}{GTL}

\maketitle
\begin{abstract}
\input{sections/0_abstract}

\end{abstract}

\input{sections/1_intro}

\input{sections/2_related_work}
\input{sections/3_probing}
\input{sections/4_method}

\input{sections/6_conclusion}

\bibliography{bib/anthology, bib/custom, bib/references}

\appendix
\input{sections/A_training_details}

\input{sections/B_grad_analysis}

\end{document}

%% file: sections/0_abstract.tex
Large Language Models (LLMs) trained on web-scale text corpora have been shown to capture world knowledge in their parameters. However, the mechanism by which language models store different types of knowledge is poorly understood. In this work, we examine two types of knowledge relating to temporally sensitive entities and demonstrate that each type is localized to different sets of parameters within the LLMs. We hypothesize that the lack of consideration of the locality of knowledge in existing continual learning methods contributes to both: the failed uptake of new information, and catastrophic forgetting of previously learned information. We observe that sequences containing references to updated and newly mentioned entities exhibit larger gradient norms in a subset of layers.  We demonstrate that targeting parameter updates to these relevant layers can improve the performance of continually pretraining on language containing temporal drift.

%% file: sections/1_intro.tex
\section{Introduction}
Pretraining over diverse datasets has been shown to encode world knowledge in the parameters of large language models (LLMs) \cite{petroni2019language, roberts2020much, gueta2023knowledge} %, xu2024knowledge}
from massive static web-scale datasets. However, these models are normally trained on large static text corpora which do not reflect changes in world knowledge or language usage that occur after the initial data collection. In practice language models are deployed in dynamic real-world settings, and their learned knowledge becomes stale over time \cite{lazaridou2021mind, luu2022time, dhingra2022time, yao2022wild, nylund2023time, cheang2023can}; %In practice language models undergo temporal degradation when deployed in real-world settings as these changes occur, 
% real-world settings where these changes result in the learned knowledge becoming stale and out-of-date; 
the temporal degradation can be evaluated according to intrinsic measures such as perplexity, or extrinsic downstream performance (e.g. question answering).

\begin{figure}
\centering
\includegraphics[width=0.5\textwidth]{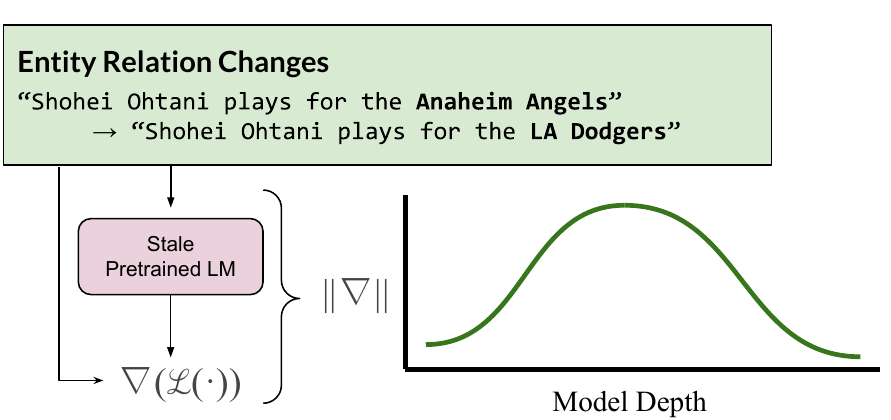}
% \vspace{-1em}
\caption{\small 
    When continually pretraining on sequences with updated and newly mentioned entities, certain layers consistently observe larger gradient norms.
% The NLL loss gradients of updated entities and newly mentioned entities observe characteristic patterns of layers with larger gradient norms.
}
\label{fig:figure_1}
\vspace{-1em}
\end{figure}

Incrementally training language models on streams of data has been explored as a method to mitigate temporal performance degradation without incurring the heavy computational and environmental costs of retraining models on large pretraining corpora \cite{jang2021towards, jang2022temporalwiki, lin2022continual, gururangan2020don}.  
However, naive online training on these data streams is known to: induce hallucinations in language generations \cite{kang2024unfamiliar}, fail in the uptake of new information \cite{Onoe2023CanLL,hu2023meta}, and catastrophically forget previously learned information \cite{zhu2020modifying}. 

To address these problems, recent work has applied continual learning and online learning methods to adapting large language models to streams of documents \cite{loureiro2022timelms, scialom2022fine, jang2022temporalwiki}. While continual learning methods have been shown to mitigate temporal performance degradations, the mechanisms by which neural language models store and update information are not well understood.
%, lin2022continual, jin2022lifelong}.

\begin{table*}[ht]
\centering
\scriptsize
    \begin{tabularx}{\textwidth}{llll}
    \toprule
    Dataset   & Year &  Example & Answer \\
    \midrule
    TempLAMA       & 2020 & \textbf{Joe Biden}  holds the position of \_\_ . &  President-elect.of the United States       \\
                                 & 2021 & \textbf{Joe Biden} holds the position of  \_\_ . &  President of the United States\\
    \midrule
    \makecell{\vspace{0.25em} \\Entity Cloze \\ By Date (ECBD)} 
        & 2020 & \makecell{
                    The Congressional Budget Office provided  a score 
                    for  the\\ \textbf{CARES Act}  on April 16, 2020 estimating it would \_\_.} 
                    &    increase federal deficits. \\
       & 2021  & \makecell{
                    On August 14, when\textbf{ Hurricane Grace} entered\\ the Caribbean, 
                    a tropical storm watch was issued for  \_\_.} 
                    &  the entire coast of Haiti.\\
    \bottomrule
    \end{tabularx}
    \caption{Examples from TempLAMA and ECBD probing tasks. The temporally sensitive entity is \textbf{bolded}.  }
    \label{tab:probe_datasets}
    \vspace{-1.5em}
\end{table*}

% As one potential solution, large-scale self-supervised continual pretraining has been shown to improve performance when training on a sequence of natural language domains \cite{gururangan2020don}, but these methods often fail to acquire new knowledge \cite{hu2023meta, }. %when \todo{Add failure modes of continual pretraining} \cite{hu2023meta, Onoe2023CanLL}.

In this work, we consider a real-world setting for continual language learning, that of temporal language drift, and probe the performance of language models on two types of entity relationships which exhibit temporal degradation: (1) acquisition of information about new entities, and (2) updating relationships between existing entities. We hypothesize that the poor performance of existing continual learning methods on these forms of entity relationship shift can be in part attributed to a misalignment in the autoregressive language modeling pretraining objective and the optimal parameter updates required to acquire new information or update existing knowledge. %\todo{one other possible descriptor: ``new'' information doesn't conflict with information in the stale data, while updates probably do}

To characterize this misalignment, we compare the gradient updates observed when training language models to predict knowledge intensive salient entity spans, with the gradient updates observed from standard continual pretraining. We observe that for the gradient updates for predicting knowledge intensive salient spans, observe high values in distinct groups of layers based on the type of entity relationship presented in the sequence (see Fig.~\ref{fig:figure_1}). 
Based on these observations, we propose new methods for aligning the gradient updates during continual pretraining to better align with these layers which exhibit high gradient norms. Through empirical study, we show that the observed characteristic gradient patterns occur across autoregressive, transformer language models of various of sizes; and we demonstrate the efficacy of our proposed method through performance improvements on knowledge probing tasks when applied on top of existing continual learning methods in pretraining.  %for continual pretraining

%% file: sections/2_related_work.tex
\vspace{-0.5em}
\section{Related Work}
\label{sec:related_work}
\vspace{-0.5em}

\paragraph{Continual Pretraining of Language Models. }
Continued pretraining of models on the target distribution is often used to adapt a generically pretrained language model from its source to its target setting to update factual knowledge or to adapt to new language domains \cite{lin2022continual, jin2022lifelong, wu2024continual}. However, standard finetuning techniques can result in catastrophic forgetting of previously learned tasks and the loss of the pretrained models generalization capabilities due to distortion of the underlying features and lack of regularization \cite{kumar2022fine}. 

As a mitigation for forgetting, it is common to apply regularizers or constraints on the gradient descent updates such as: gradient projection, example-replay, loss rescaling, or introduction of additional parameters for the target domain \cite{cossu2022continual,saha2021gradient,farajtabar2020orthogonal}. While continual pretraining is commonly used in the adaptation to a sequence of domains \cite{gururangan2020don, yildiz2024investigating}, recent work is only beginning to explore its use in the adaptation to changing temporal knowledge which can often exhibit finer-grained changes \cite{jang2021towards, jang2022temporalwiki, nylund2023time}. 

\paragraph{Knowledge Localization and Model Editing.} 
Another method to adjust the information contained within large pretrained models is knowledge editing, in which specific factual relations are injected or manipulated by performing causal traces of activations to identify where a model stored knowledge necessary for prediction \cite{de2021editing, meng2022locating, meng2022memit}. However, these methods exhibit high per-edit computational costs and fail to large number of edits \cite{gupta2024model}, which can become necessary when updating models over larger corpora or repeatedly over time.

%% file: sections/3_probing.tex
\section{Knowledge Probing Using Salient Span} % Prediction}
\label{lab:probing}
We probe language models using the task of salient span prediction, which has previously shown success as a pretraining objective for knowledge-intensive tasks such as closed-book question answering \cite{cole-etal-2023-salient, guu2020retrieval}. In salient span prediction, a model is provided with a sequence and tasked with completing a masked slot corresponding to a named entity or noun phrase. Specifically, we examine language models on probing tasks for temporal entity knowledge in which the masked sequence corresponds: (1) to an update or change to an existing temporally sensitive entities; (2) to a mention of emerging new entities that were not previously seen during pretraining.

\begin{figure*}
    \centering
    \begin{subfigure}[t]{0.4\textwidth}
        \centering
        \subfloat{\includegraphics[width=0.4\textwidth]{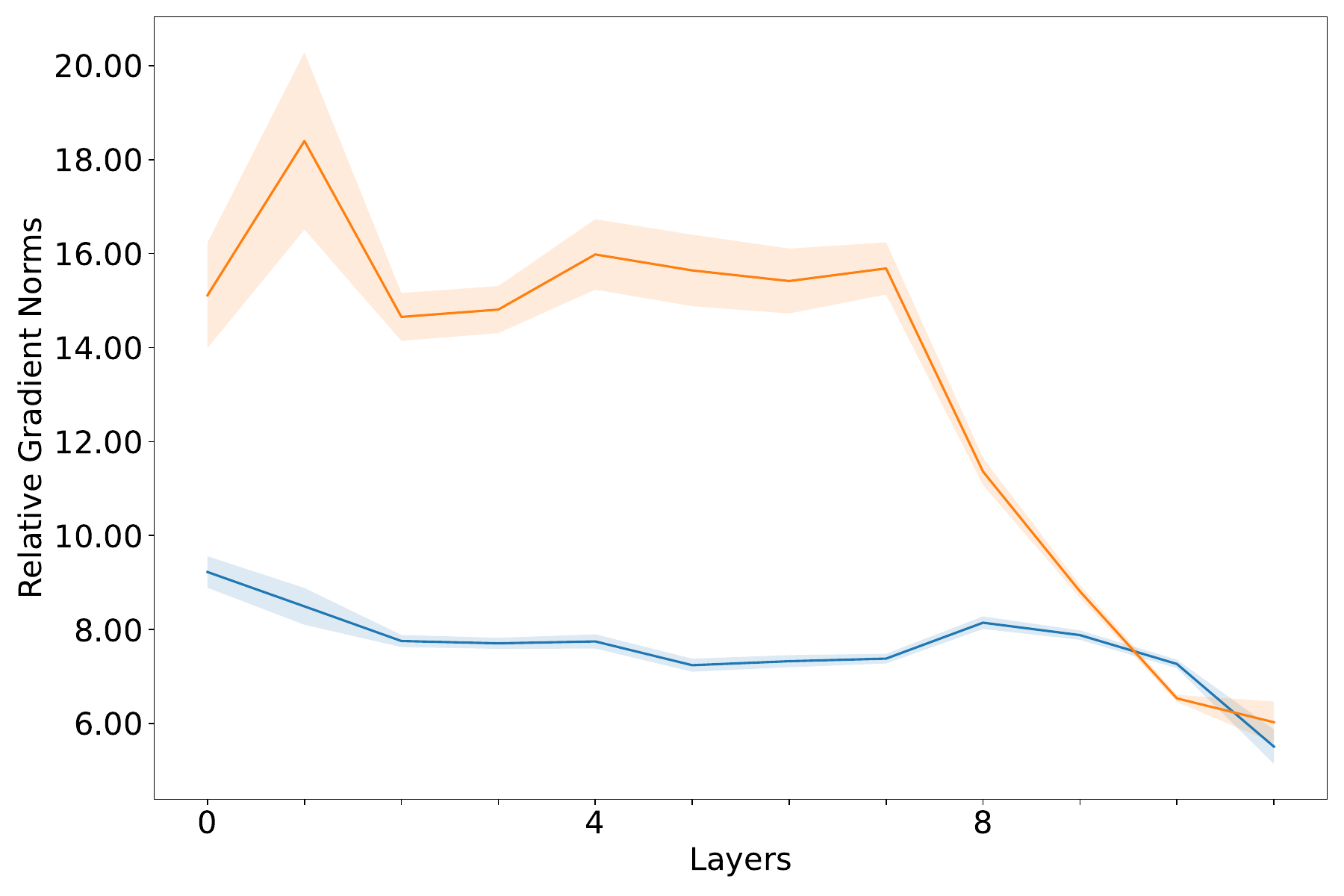}}
        \hspace{2em}
        \subfloat{\includegraphics[width=0.4\textwidth]{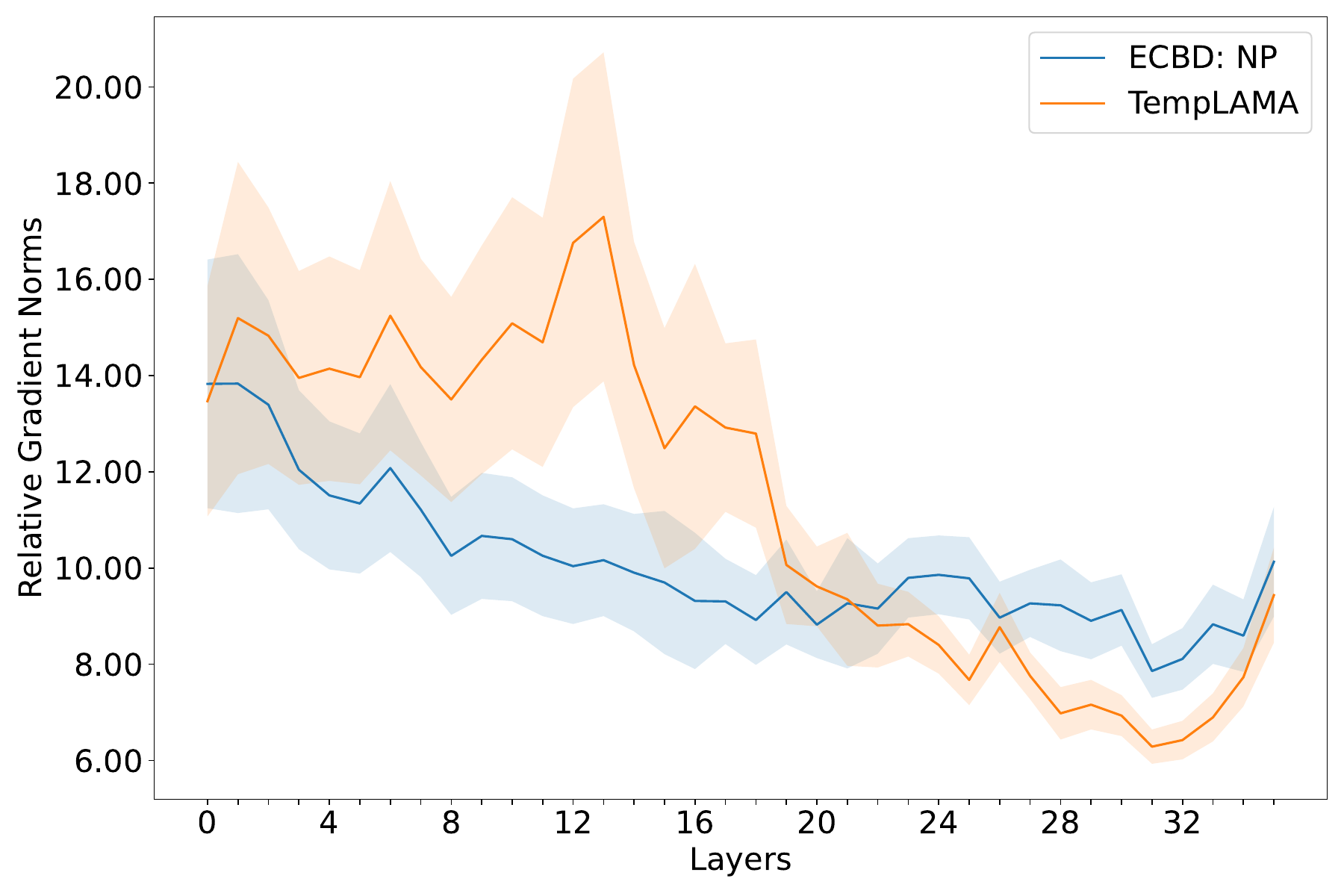}}        
    \end{subfigure}
    \begin{subfigure}[t]{0.4\textwidth}
        \centering
        \subfloat{\includegraphics[width=0.4\textwidth]{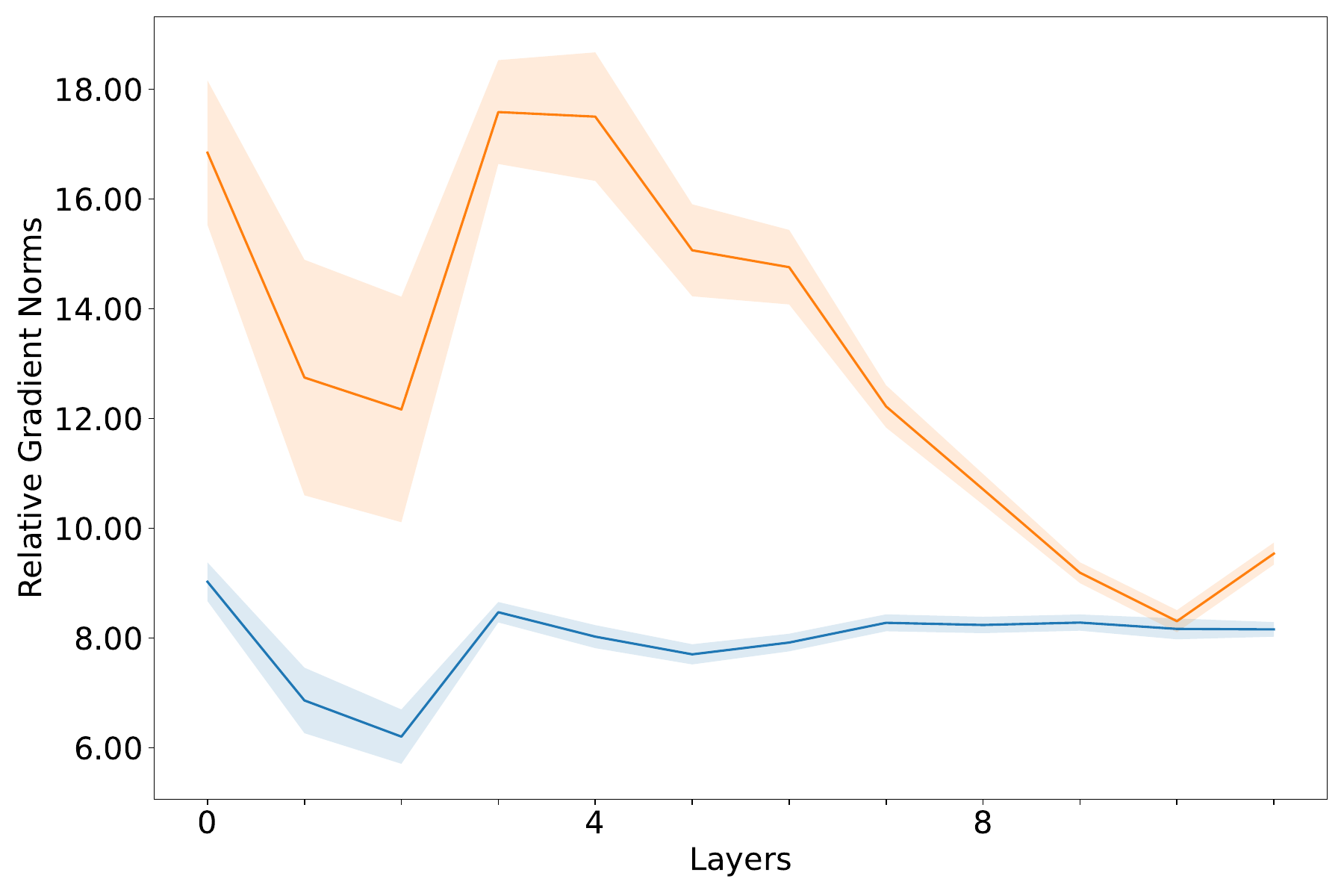}}
        \hspace{2em}
        \subfloat{\includegraphics[width=0.4\textwidth]{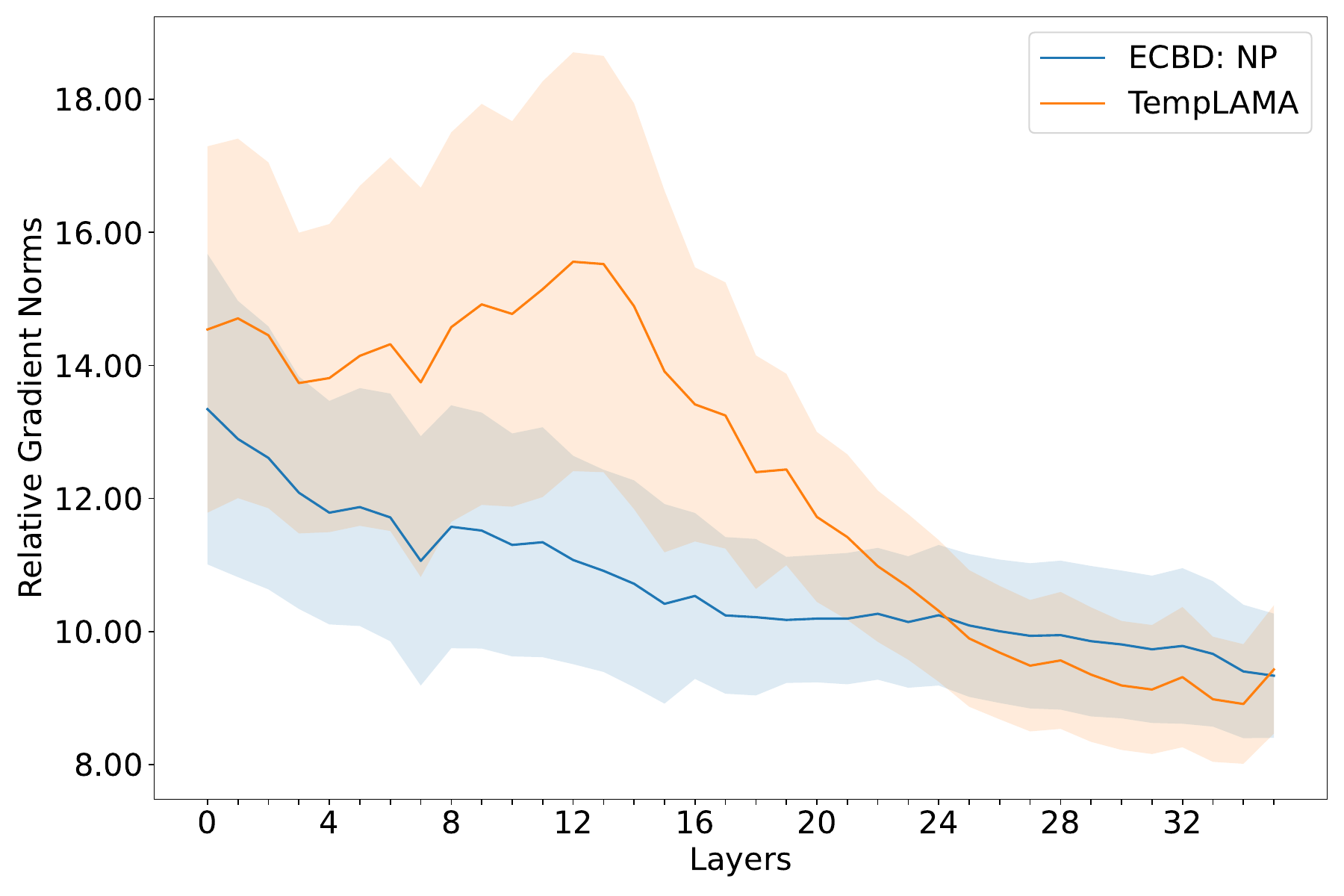}}
    \end{subfigure}
    \caption{
        Relative gradient norms for the salient spans in ECBD and TempLAMA for the GPT-2 Base (110M; Left-hand side), and GPT-2 Large (770M; Right-hand side), models. Norms for attention (Top) and norms for MLP (Bottom) are depicted separately. Rradient norms of salient spans are 4 to 15x larger than those of the full sequence. 
    }
    \label{fig:gradient_analysis}
\end{figure*}

\subsection{Probing Datasets}
\label{sec:probing_datasets}
We study these  using the Dynamic TempLAMA \cite{dhingra2022time} and the Entity Cloze By Date \cite{onoe2022entity} diagnostic datasets, respectively.  Examples can be found in Table \ref{tab:probe_datasets}. 

\paragraph*{Dynamic TempLAMA} contains cloze queries consisting of subject-object relations in which the correct answer corresponds to objects that have changed over time. 
% Examples are generated from natural language templates based relations extracted from Wikipedia metadata, and are generated sequentially for three month periods. For our analysis, we examine splits for each year from 2019 to 2021. 
Although the answer may change over time, the referenced subject in each example may have been mentioned in both the seen data (i.e. initial pretraining corpus) and unseen data (i.e. continual pretraining corpus. Thus, we use this dataset to evaluate the ability of continual learning techniques to \textit{update} existing knowledge. 

\paragraph*{Entity Cloze By Date} contains cloze queries where the salient spans correspond to noun-phrases (ECBD-NP) referring to newly emerging entities that are not seen prior to specified cutoff dates. As the entity was not seen in initial pretraining but may have been mentioned in the subsequent continual pretraining, we use the ECBD-NP dataset to evaluate the effectiveness of a continual learning method in \textit{knowledge acquisition.}
% in the pretraining corpus; which can be used to evaluate the effectiveness of continual learning methods in \textit{knowledge acquisition}. Examples are grouped by year, according to the first time of mention. 

Additionally, we evaluate on the ECBD-Popular split in which the salient spans reference entities that exists in all splits. As the ECBD-Popular split references static information that was seen in both the intial and continual pretraining data, we use the ECBD-Popular split to evaluate \textit{catastrophic forgetting} the \textit{retention of previously learned knowledge}.

% ECBD was constructed from a more recent 2022 snapshot of Wikipedia; To avoid data contamination, we  deduplicate data 

% For both datasets, we evaluate the extent to which the language model contains the relevant temporal knowledge by computing the mean and median per token-normalized perplexity over the target span as in prior work \cite{onoe2022entity}. 

% We also report the exact match performance in generating the salient span.
% Include note on dataset statistics.

\subsection{Models}
We examine decoder-only transformer language models of various sizes, specifically: GPT 2-Base (110M parameters) and GPT-2 Large (770M parameters); with additional analysis on GPT-Neo (1.3B parameters) in Appendix \ref{fig:neo-relative-grad}. To evaluate the perplexity of each of these models, we provide the example context of each example up to the salient span and compute the perplexity over the salient span as in \cite{onoe2022entity, Onoe2023CanLL}.

To align each model with Wikipedia-based knowledge contained in the probing tasks, we perform domain adaptive pretraining on snapshots of Wikipedia retrieved prior to the pretraining data cutoffs for each model to prevent data contamination. Speicifically, we perform initial pretraining of GPT-2 models on Wikipedia snapshots from January 2019, and of GPT-Neo on January 2020.
% and T5 pretraining on snapshots from January 2020. 
% As with the original pretraining objective of each model, we train GPT models with an autoregressive left-to-right language modeling and T5 models with sequence-to-sequence span prediction. 
% We evaluate models on yearly subsets to examine the performance of model on changes in relationships and emerging entities.

% Removing as we don't have performance numbers for the probing tasks; will add back later for rebuttal
% Additionally, we examine the sequence-to-sequence transformers, specifically: T5-Base (220M parameters), and T5-Large (770M parameters).

\subsection{Probing Model Response to Salient Spans}
% Compare with localization in model editing papers -- Hasse, 23 shows that causal tracing (reperesentation denoising) is insufficient in determining which layer to edit
% TODO: Add a motivating statement of how, where, why we plan to probe models with gradient norms. Find the paper mentioend by Sang on embedding layer knowledge; 
We hypothesize that the target parameters and gradient updates relevant for learning the entity relationship knowledge previously described in Section \ref{sec:probing_datasets} differs from those observed during autogressive continual pretraining. Based on this hypothesis, we analyze the per-layer gradient norms for examples which reflect the target form of knowledge. 

% TODO phrasing; Cite the surgical finetuning paper 

% Most of this can probably be left as technical detail in the appendix
To identify critical portions of the model, we compare the relative gradient norms for salients span prediction on the knowledge probing tasks with the gradient norms of randomly sampled autoregressive pretraining examples. Precisely, we provide the autoregressive language model with the left context preceding the salient span and compute the parameter gradient with respect to the loss, averaged over each token in the target span.  We then aggregate the gradients according to their respective transformer block's component attention and MLP layers, and compute the L2-Norm of the gradients for each layer. We then normalize these per-layer norms with the average per-layer gradients for 2000 examples from the 2019 Wikipedia snapshot over the full sequence.

% We hypothesize that the  portions of the model are responsible for different forms of knowledge can be identified by tracing the gradient norm of examples which reflect the target form of knowledge. % TODO phrasing; Cite the surgical finetuning paper

% We 
% Beginning with a domain-adapted model pretrained on a snapshot of Wikipedia from 2019, 

For the ECBD probing dataset, we examine the gradients for the salient span corresponding to the noun phrase related to the target entity, which we refer to ECBD-NP. For the TempLAMA dataset, we examine the loss gradient with respect to the object noun phrase. % As the entity span occurs at the start of the probe sequence, so its loss is confounded by the high certainty around its unconditioned generation. 

 % Consider explaining why we are probing knowledge for already seen data

% To understand how the updates for 

% Consider an example of the form 
%  $\nabla_\theta \left( \frac{1}{j - i} \sum_{t \in [i,j]} \mathcal{L}_{CE}(f_\theta(x_i), y_i)  \right)$

% \todo{xlimit to 0 and 36; Bigger font and weight on line; Vertically align plots by model}
% \begin{figure*}
%     \centering
%     \subfloat{\includegraphics[width=0.33\textwidth]{figures/gpt2_attn.pdf}}
%     \subfloat{\includegraphics[width=0.33\textwidth]{figures/gpt2-large_attn.pdf}}
   
%     \subfloat{\includegraphics[width=0.33\textwidth]{figures/gpt2_mlp.pdf}}
%     \subfloat{\includegraphics[width=0.33\textwidth]{figures/gpt2-large_mlp.pdf}}
%     \caption{Relative gradient norms for the salient spans in ECBD and TempLAMA for the GPT-2 Base (110M), \textit{left}, and GPT-2 Large (770M), \textit{right}, models. Norms for attention are depicted \textit{top}; norms for MLP are depicted \textit{bottom}. Gradient norms of salient spans are 4 to 15x larger than those of the full sequence. }
%     \label{fig:gradient_analysis}
% \end{figure*}

% TODO: expand, formal notation -- not clear atm
% TODO: include absolute gradient norms -- maybe appendi
In Figure \ref{fig:gradient_analysis}, we observe that the gradient norms for salient spans are consistently 4 to 15x higher than the gradient norms of randomly sampled pretraining examples for all layers in both GPT2-Base and Large. Additionally, we observe that the relative gradient norms for these salient spans observe a distinct profile in which there is large magnitude in the early and middle layers, and that the relative gradient norms are larger in the attention layers than in the MLP layers. 

%% file: sections/4_method.tex
\section{Gradient Localized Continual Pretraining}
\label{lab:grad_analysis}
Ideally, standard autoregressive pretraining of a language model on a changing stream of data would be sufficient to update a model to capture the relevant changes in knowledge. However, recent work  % add other cites
has demonstrated that current methods for continual learning often suffer from both catastrophic forgetting and a failure to uptake new knowledge even when it is explicitly contained in the training corpus \cite{hu2023meta, kang2024unfamiliar}.
Based on our observations from \S\ref{lab:probing}, we hypothesize that one cause of failed transfer is due to a misalignment of the gradients from the NLL objective function with the desired update based on the information content of the data observed during continual pretraining.

% Motivated by \cite{lee2022surgical} , which demonstrated that gradient norm can be used to adaptively set learning rates for different blocks in the model, we propose using the characteristic gradient patterns observed in \ref{sec:grad_analysis} to guide optimization during continual pretraining. 
% Previous work \cite{zhu2020modifying, houlsby2019parameter} has shown that constraining updates to a subset of early and late layers finetuning is sufficient to improve language model performance during finetuning on factual edits, even beyond that of full fine tuning. However, these previous approaches provide limited guidance on how to identify which layers to target for different tasks. 

We propose a method to improve the acquisition of entity knowledge by amplifying updates to the layers that are relevant to the learning of salient entity spans. To identify relevant layers, we compute the relative gradient norm for each layer $i$ as: the ratio between the gradient norm $\Tilde{\nabla_i}$ in the layer $i$ for knowledge intensive salient prediction on data sampled from the validation set of the TempLAMA diagnostic dataset, and the gradient norm for autoregressive pretraining on randomly sampled data from the continual pretraining data stream: 
\begin{align} 
\Tilde{\nabla_i} = 
\frac{
||\nabla_i   \mathcal{L}(M_\theta, (x,y)_{\text{TempLAMA}})||
}{
||\nabla_i   \mathcal{L}(M_\theta, (x,y)_{\text{PT}}) ||
}
\end{align}

We propose two methods to improve knowledge uptake by aligning gradient updates during continual pretraining. For relevant salient spans from the TempLAMA diagnostic dataset, we construct a profile of the relative gradient norms with respect to the gradients for randomly sampled pretraining sequences. We then adjust the learning rates for layers in this profile to increase the updates to layers with large relative gradient norms. We refer to our methods as Traced Gradient Layers (TGL).

\paragraph{Selecting Trainable Layers for Pretraining with Relative Gradient Norm}

We consider a simple approach to target continual pretraining updates to layers with high relative gradient norm, by only updating parameters where the relative gradient norm on the TempLAMA diagnostic dataset exceed the mean relative gradient norm of all layers -- we refer to this method which freezes parameters as TGL + FP. In the case of the GPT-2 architecture, we separate the model into its component MLP and attention layers, then compute the relative gradient norm for each layer as the ratio between the average gradient norm computed over salient spans from the TempLAMA dataset and the gradients for examples from the continual pretraining corpus.  Precisely, we freeze a parameter group for layer $i$ if $\Tilde{\nabla_i} < \frac{1}{\text{No. Layers}}(\sum_{k \in \text{Layers} } \Tilde{\nabla_k})$. 
% Parameter Efficient, relies on setting a threshold
% For each probing task, we determine whether to freeze or train a layer by examining which layers exhibit high relative gradient norm. We compare the  to the gradients computed from a randomly sampled set of 10,000 pretraining examples. 
% How/why do we arrive at 10k pretraining examples; which pretraining split are they sampled from?
% We select layers where the relative gradient norm  exceeds the mean over all layers. For each MLP and self-attention layer in the model, we compute the relative gradient norm with resepect to the pretraining 

% In thse experiments, we allow the embedding layer to remain unfrozen -- in appendix B, we observe that 
% Add a statement on what is required to determine layer selection and learning rates -- this is currently/will be mentioned in Appendix B;
\paragraph{Per-Layer Adaptive Learning Rates from Relative Gradient Norm}
Rather than using relative gradient norm as a hard threshold to determine which layers to update, we instead consider an adaptive approach in which we set the learning rate for layers to scale with the magnitude of the relative gradient norm; we refer to this method as TGL + ALR. We scale the per-layer learning rate for layer $i$ as :
$\eta_i = \eta \frac{\Tilde{\nabla_i}}{\max_{i \in {\text{Layers}}}(\Tilde\nabla_k)} $

% Adaptive no threshold required, requires training all parameters
% Rather than using relative gradient norms as a threshold for selecting layers for fine-tuning, we instead use the relative gradient norm to adaptively set per-layer learning rates that prioritize learning layers with high gradient norm. 
% For each MLP and attention layer, we set its learning rate as:
% $ ||\nabla_{\text{Relative}}(Span)||  = 
% \frac{||\nabla_\theta_i(\mathcal{L}_{\text{CE, Span}}) ||} {||\nabla_\theta_i(\mathcal{L}_{\text{CE, Sequence})}||} $ 
% $\eta_{\theta_i} = 
%     \eta_{\text{Base}} \frac{||\nabla_{\text{Relative}}||}{\max(\nabla)}$

% \subsection{Method 3: Relative Gradient Norm Relative to Paraameter Norm}  -- 
% \vspace{-5}
\section{Training and Dataset Details}
To perform domain adaptive pretraining, we sample and preprocess a snapshot of Wikipedia from January 2019 using Wikiextractor. For continual pretraining, we follow the methodology of \cite{jang2022temporalwiki} to collect snapshots of Wikipedia from each of the subsequent years until 2022 and filter each corpus to contain the edits to Wikipedia made in the intervening year, consisting of new articles and sentences within existing articles that were edited between succeeding snapshots.

\subsection{Baselines}
% We provide full details on the training datasets and hyperparameters in the Appendix. % Appendices \ref{apx:datasets, apx:train_details}.
We compare the performance of our proposed continual pretraining method with existing approaches from continual learning. We consider vanilla continual pretraining in which we update all parameters; a parameter-expansion method LoRA \cite{hu2021lora}, which introduces additional trainable low rank adapters to the self-attention layers; a replay-based method MixReview \cite{he-etal-2021-analyzing}, which randomly mixes previously seen pretraining data alongside current data; and the regularization-based method RecAdam \cite{chen2020recall}, which imposes a quadratic penalty on the norm of the parameter update.

Initial domain adaptive pretraining is performed on a the complete Wikipedia snapshot for 4 epochs with a global batch size of 64, or approximately 500,000 training iterations. Models are trained using the Adam optimizer with weight decay and a linear warmup schedule over 10\% of examples and a linear decay with a max learning rate of 1E-4.

During continual pretraining, the model is trained for one epoch on the Wikipedia edits for the subsequent year. For the MixReview method, unedited articles are added Wikipedia edits corpus at a 2:1 ratio. We train LoRA adapters with a hidden rank of 64 dimensions.

\begin{table}[t!]
    \scriptsize
    \centering
    \begin{tabular}{lccc}
        \toprule
        \textbf{Evaluation Set: 2020}       & \textbf{ECBD Pop.} & \textbf{ECBD NP} & \textbf{TempLAMA} \\ \midrule
        Pretrain           & 40.99              & 47.44            & 81.92             \\ 
        Domain Pretrain    & 30.90              & 41.39            & 62.99             \\ \midrule
        Continual Pretrain & 34.79              & \textbf{43.97}            & 56.72             \\
        + TGL with FP      & \textbf{34.13}     & 44.20            & \textbf{55.19}    \\ \midrule
        LoRA: 64D, Attn    & 31.94              & 41.40            & 57.21             \\
        + TGL with FP      & \textbf{30.28}     & \textbf{41.05}   & \textbf{56.32}    \\ \midrule
        MixReview          & 28.70              & \textbf{37.34}   & 67.64             \\
        + TGL with FP      & \textbf{28.24}     & 37.77            & \textbf{60.05}    \\ \midrule
        RecAdam            & 34.78              & 43.92            & 57.34             \\
        + TGL with FP      & \textbf{33.56}     & \textbf{43.41}   & \textbf{54.75}    \\ \bottomrule
    \end{tabular}
    \caption{TGL with frozen layers improves performance of GPT2-Large (770M) during continual pretraining. }
    \vspace{-2em}
    \label{tab:gpt2-large}
\end{table}

\begin{table}[t!]
\scriptsize \centering
    \begin{tabular}{lccc}
        \toprule \textbf{Evaluation Set: 2020 }  & \textbf{ECBD Pop.} & \textbf{ECBD NP} & \textbf{TempLAMA} \\
        \midrule \midrule
        Pretrain              & 78.61           & 80.04  & 162.54                   \\ \midrule
        Domain Pretrain       & 55.26 & 62.59  & 80.51                            \\ \midrule
        Continual Pretrain    & 64.13           & 72.42           & 83.39           \\
        + TGL with ALR        & \textbf{57.62}  & \textbf{64.83}  & 77.58           \\
        + TGL with FP         & 57.75           & 65.08           & \textbf{74.55}  \\ \midrule
        MixReview             & 54.10           & 61.54           & 82.16           \\
        + TGL with ALR        & 53.50           & \textbf{61.01}  & 77.04           \\
        + TGL with FP         & \textbf{53.48}           & 61.48           & \textbf{76.35}  \\ \midrule
        LoRA                  & \textbf{55.77}           & \textbf{65.56}          & 80.11           \\
        + TGL with ALR        & 57.75           & 69.44           & \textbf{78.40}           \\
        + TGL with FP         & 58.09  & 67.62           & 78.77           \\
        \midrule
        RecAdam               & 57.55           & \textbf{64.60}           & 76.67           \\
        + TGL with ALR        & \textbf{57.52}  & 64.77           & 77.32           \\
        + TGL with FP         & 57.55           & 64.89           & \textbf{74.88}  \\ 
        \midrule \midrule
        \textbf{Evaluation Set: 2021} & \textbf{ECBD Pop.} & \textbf{ECBD NP} & \textbf{TempLAMA} \\ \midrule \midrule
        Pretrain              & 78.61          & 98.47          & 167.23         \\ \midrule
        Domain Pretrain     & 55.26          & 66.16          & 82.60          \\ \midrule
        Continual Pretrain    & 67.18          & 77.70          & 86.34          \\
        + TGL with ALR        & 57.91          & \textbf{63.45} & 78.85          \\
        + TGL with FP         & \textbf{57.83} & 63.55          & \textbf{74.88} \\ \midrule
        MixReview             & \textbf{51.96} & 57.69          & 81.88          \\
        + TGL with ALR        & 53.42          & 59.60          & \textbf{78.75}          \\
        + TGL with FP         & 52.81          & \textbf{58.31} & 79.17          \\ \midrule
        LoRA                  & 58.07 & 66.89          & \textbf{76.78}          \\
        + TGL with ALR        & \textbf{58.06 }         & 69.17          & 79.03          \\
        + TGL with FP         & 58.39          & \textbf{66.31} & 78.19          \\ \midrule
        RecAdam               & 64.42          & 73.34          & 92.26          \\
        + TGL with ALR        & 57.72          & \textbf{63.53}          & 78.39          \\
        + TGL with FP         & \textbf{57.69}          & 63.60          & \textbf{75.21} \\ \midrule \midrule
    \end{tabular}
    \caption{
    Traced Gradient Layers (TGL) can be applied on top of existing continual pretraining methods by applying per-layer adaptive learning rates (ALR) or frozen parameters (FP) to improve performance (perplexity of the slot) of existing continual learning methods. 
    }
    \vspace{-2em}
    \label{tab:gpt2-base}
\end{table}

\subsection{Evaluating TGL for Continual PT}
To evaluate the performance of TGL+FP and TGL+AR, we incrementally train the domain-adapted language model on the subsequent set of Wikipedia revisions for the years of 2020 and 2021. We then probe the continually pretrained model after each updating on new year of Wikipedia revisions using the corresponding temporally delineated split from the ECBD-NP and TempLAMA test datasets \ref{sec:probing_datasets}. To evaluate whether either TGL method leads to catastrophic forgetting, we also report performance on ECBD-Popular, which contains sequences referring to entities common in all years including entities previously seen during initial pretraining. 

% Please add the following required packages to your document preamble:
% \usepackage[table,xcdraw]{xcolor}
% Beamer presentation requires \usepackage{colortbl} instead of \usepackage[table,xcdraw]{xcolor}

% \vspace{-10}

In Table \ref{tab:gpt2-base}, we report the perplexities of the continually pretrained model on the 2020 and 2021 test splits with the GPT-2 Base (110M) model. Relative to the domain-adapted pretrained initialization, we observe that all continual learning baselines exhibit performance tradeoffs in which performance either improves on the probe tasks for recognizing new entities (ECBD-NP) \textit{or} improves on updating entity relations (TempLAMA). 

When applying TGL methods on top of continual learning methods, we see that it is possible to avoid catastrophic forgetting as we observe decreases in probing task perplexity relative to the continual learning baselines.   % Additionally, as the initial pretrained  yield model becomes out of date the 
In Table \ref{tab:gpt2-large}, we scale our experiments to the GPT-2 Large (770M) model and observe that the improvements from localized gradient updates extend to continual pretraining for the larger model. 

\definecolor{mygreen}{RGB}{0, 153, 51}
\newcommand{\DA}{\textcolor{mygreen}{$\downarrow$}}
\newcommand{\UA}{\textcolor{red}{$\uparrow$}}
\newcommand{\prop}{\hspace{1em} + TGL}

%% file: sections/6_conclusion.tex
\section{Conclusion}
In this work, we conduct an analysis of the gradient updates observed during knowledge intensive salient span prediction and autoregressive language modeling, and observe characteristic differences in the layer-wise norms for each objective. Based on this observation, we proposed Traced Gradient Layers (TGL) a method for identifying relevant layers to target during continual pretraining of language models. We observe that our proposed approach improve language model performance on tasks probing for entity and relational knowledge; without the need for fine-grained annotations.

% % Future Directions
% % 1. Other types of online & continual learning, domain expansion: 
% %     A. Continual learning code LMs
% %     B. Expansion and Analysis of Experts Across Language Domains
% %     C. PT -> Instruction Tuning -> Domain Training
% %     D. Temporal adaptation and expression of previously learned knowledge from pretraining data
% % % 2. Design of Architectures that Mitigate Interference: MoE's, parameter expansion
% % % 3. Analysis of Training Dynamics Across Other Types of Models
% % %     A. Multimodal, Robotics
% % %     B. What regularizations enable consistent learning across modalities and domains;
% % % 4. Question as to how much this is related to autogressive objectives

\section*{Acknowledgements}
{ % \small
    The authors would like to thank Sanket Vaibhav Mehta for helpful discussions, as well as Clara Na and Jeremiah Milbauer for manuscript feedback. This work was supported in part by funding from the National Science Foundation Graduate Research Fellowship Program under Grant No. DGE2140739, and by DSO National Laboratories.
}

\section*{Limitations and Ethical Considerations}
In our work, we observe that per-layer gradient norms can be utilized as an informative indicator for identifying layers to train during continual pretraining on temporally changing data. Although perplexity is a commonly used metric for evaluating language models and can often be useful in measuring the quality of a model, it is unclear whether improvements in knowledge probe perplexity transfers to downstream settings.

While the goal of our investigations is to mitigate the need for environmentally and financially prohibitive pretraining by enabling the continual learning of existing models, it is possible that reductions in the cost of pretraining may then lead more individuals and organizations to pursue large model pretraining (i.e. Jevons Paradox).

% \section*{Acknowledgements}
% This work was supported by in part by the DSO National Laboratories and the National Science Foundation Graduate Research Fellowship. The authors would like to thank Clara Na, Jeremiah Millbauer, Sanket Vaibhav Mehta, Saujas Vaduguru, Adithya Pratapa, and others for helpful discussions and feedback.

% Understanding methods for adapting models to a specific type of knowledge or information could enable a bad actor 

%  Confounds from types of entities that are changing
% We focus on the editing and acquisition of temporally sensitive data; This does not capture unlearning of relations
% Model architecture
% Weakness of LM perplexity as a metric

%% file: sections/A_training_details.tex
\appendix

\section{Licenses}
Wikipedia data, which was used to construct the TempLAMA and ECBD, the datasets we used, has a Creative Commons Attribution-ShareAlike 4.0 International License (CC BY-SA). TempLAMA is also derived from LAMA which has a CC Attribution-NonCommercial 4.0 International License (CC BY-NC 4.0), and the script for constructing it is licensed under the Apache License, Version 2.0.

Our use of the datasets is for research purposes only and aligns with the intended use.

\section{Dataset Details}
Examples from the Dynamic TempLAMA and ECBD probing and evaluation datasets are provided in Table \ref{tab:probe_datasets}.

Details on the datasets used for domain-specific and continual pretraining are provided in Table \ref{table:training_splits}.

\begin{table}[h!]
    \footnotesize
    \centering
    \begin{tabular}{llll}
        \toprule 
        Split & Date           & No. Articles & No. Tokens   \\
        \midrule
        Complete & Jan. 2019 & 7.9 Million  & 1.81 Billion \\
        Edits & Jan. 2020    & 364,235      & 268 Million  \\
        Edits & Jan. 2021    & 419,879      & 311 Million  \\
        Edits & Jan. 2022    & 425,296      & 309 Million  \\
        \bottomrule
    \end{tabular}
    \caption{Statistics on the Wikipedia corpora used for domain adaptive and continual pretraining.}
    \label{table:training_splits}
\end{table}

% Domain adaptive pretraining for the StreamingQA and WMT dataset is performed by sequentially training on complete WMT news articles from 2007 to 2019 for one epoch, or approximately [FILL ME IN] training steps. 

%% file: sections/B_grad_analysis.tex
% \section{Additional Notes on Gradient Analysis}
% \label{appx:grads}
% \subsection{Absolute Gradient Norms}
% \todo{Include figures}

\section{Gradient Profiles for GPT-Neo (1.3B)}
In addition probing the 110M and 770M parameter GPT-2 models in Section \ref{lab:probing}, we examine the gradient characteristics of the larger GPT-Neo (1.3B parameter) model.  As the GPT-Neo model was pretrained on the Pile with a data cutoff year of 2020, we conduct initial domain adaptive pretraining on a snapshot of Wikipedia from January 2020, and conduct gradient norm probes using TempLAMA and ECBD evaluation splits from 2020. 

\begin{figure}[h!]
    \centering
    \includegraphics[width=0.40\textwidth]{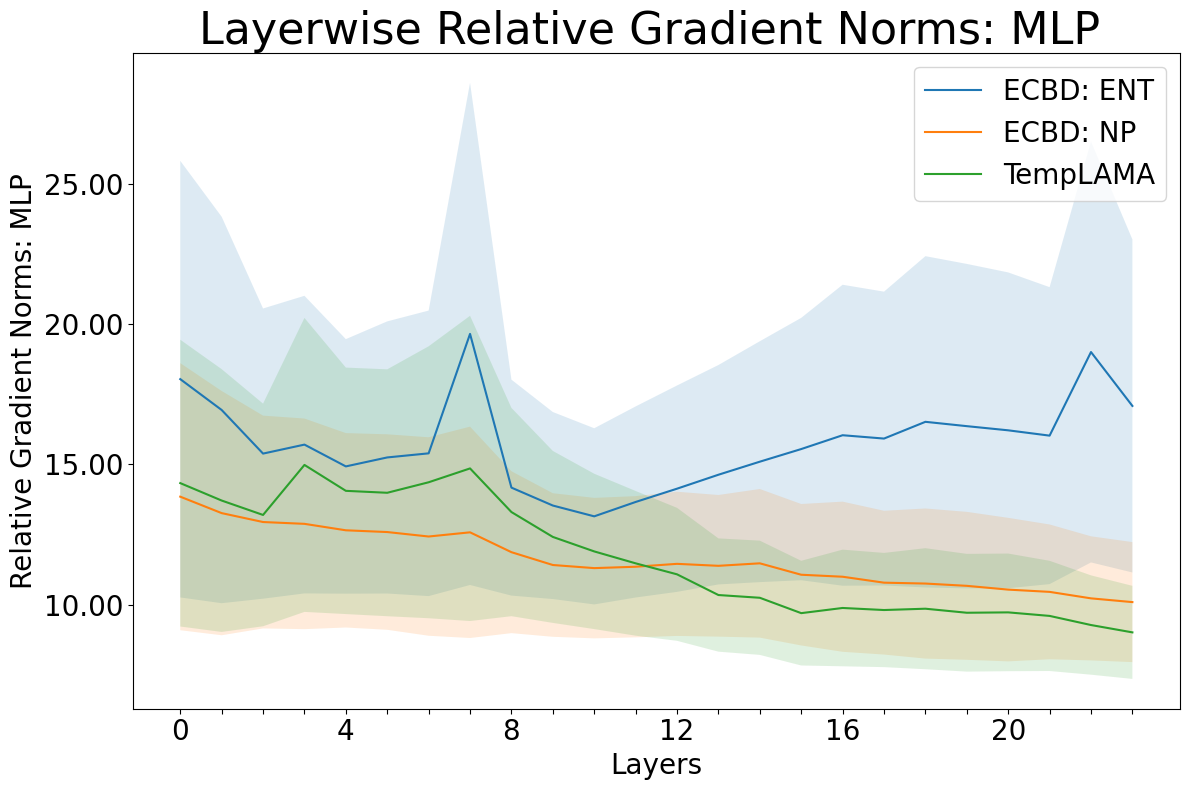}
    \includegraphics[width=0.40\textwidth]{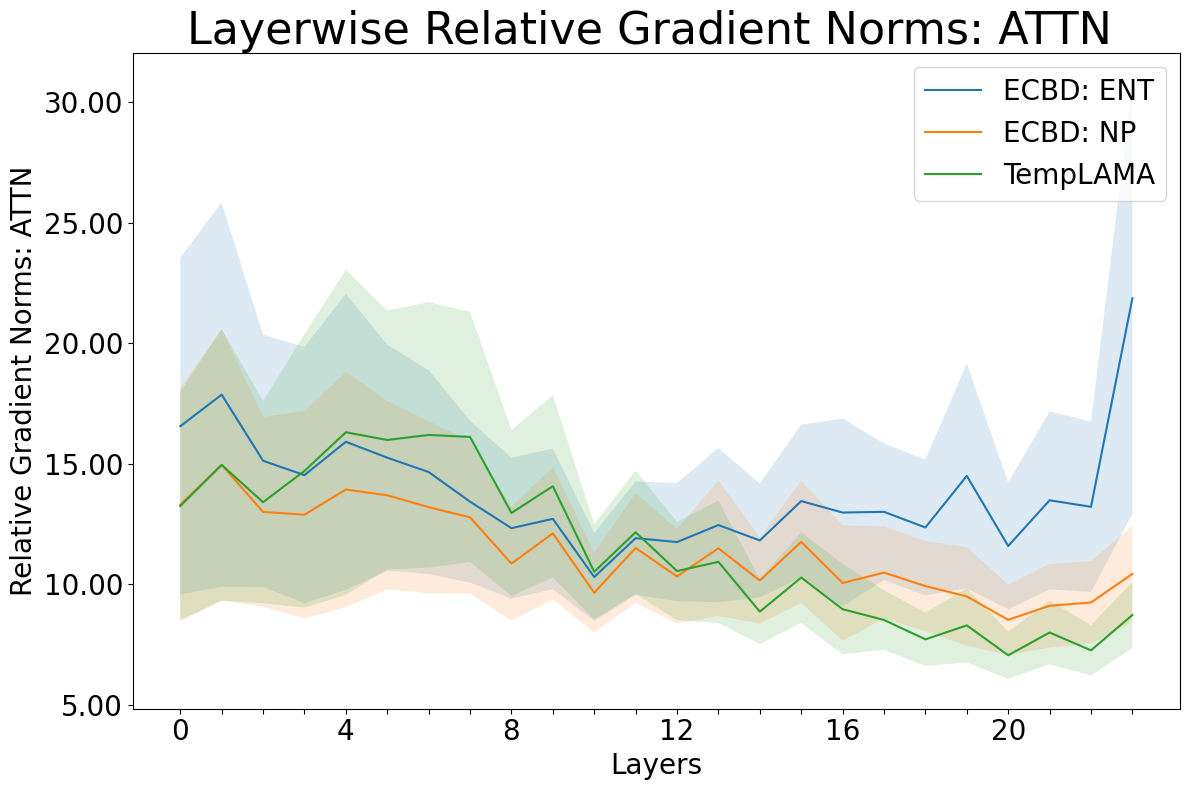}
    \caption{Relative Gradient Norms for the GPT-Neo 1.3B parameter model.}
    \label{fig:neo-relative-grad}
\end{figure}

For GPT-Neo, we observe similar characteristic gradient profiles, with increases in relative gradient norm in the first and final layers for the ECBD new entity probes (ECBD-ENT), as well as an increase in relative gradient norm in the middle layers for probes of relational changes (TempLAMA) in Figure \ref{fig:neo-relative-grad}.